# Cartesian stiffness matrix of manipulators with passive joints: analytical approach

Pashkevich A., Klimchik A., Caro S., and Chablat D.

*Abstract*—The paper focuses on stiffness matrix computation for manipulators with passive joints. It proposes both explicit analytical expressions and an efficient recursive procedure that are applicable in general case and allow obtaining the desired matrix either in analytical or numerical form. Advantages of the developed technique and its ability to produce both singular and non-singular stiffness matrices are illustrated by application examples that deal with stiffness modeling of two Stewart-Gough platforms.

## I. INTRODUCTION

IN many applications, manipulator stiffness becomes one of the most important performance measures of a robotic system. In particular, for milling, drilling and other types of machining, the stiffness defines the positioning errors due to interaction between the workpiece and the technological tool. Similarly, in industrial pick-and-place automation, the manipulator stiffness defines admissible velocity/acceleration while approaching the target point, in order to avoid undesirable displacements due to inertia forces. Other examples include medical robots, where elastic deformations of mechanical components under the task load are the primary source of positioning errors.

Numerically, this property is usually described by the stiffness matrix $\mathbf{K}_C$, which defines a linear relation between the translational/rotational displacement in Cartesian space and the static forces/torques causing this transition (assuming that all of them are small enough). The inverse of $\mathbf{K}_C$ is usually called the compliance matrix and is denoted as $\mathbf{k}_C$. As it follows from related works, for conservative systems, $\mathbf{K}_C$ is $6 \times 6$ semi-definite non-negative symmetrical matrix but in general case its structure may be non-diagonal to represent the coupling between the translation and rotation.

The problem of stiffness matrix computing for different types of manipulators is the focus of robotic experts for several decades [1-18]. The existing approaches may be roughly divided into three main groups: (i) the Finite Element Analysis (FEA) [19-23], (ii) the Matrix Structural Analysis (SMA) [24-25], and (iii) the Virtual Joint Method (VJM) [1-2,8,14,26-27]. The most accurate of them is obviously the FEA-based technique but it requires rather high computational expenses. The SMA is less computationally hard due to fairly large structural elements employed (3D flexible beams instead of numerous tiny tetrahedrals and hexahedrals of FEA) but it nevertheless is not convenient for the parametric analysis. And finally, the VJM method is the most attractive in robotic domain since it operates with an extension of the traditional rigid model that is completed by a set of compliant virtual joints (localized springs), which describe elastic properties of the links, joints and actuators. This paper contributes to the VJM technique and focuses on some particularities of the manipulators with passive joints.

For conventional serial manipulators (without passive joints), the VJM approach yields rather simple analytical presentation of the desired stiffness matrix $\mathbf{K}_C$. Relevant expression $\mathbf{K}_C = \mathbf{J}_\theta^{-T} \cdot \mathbf{K}_\theta \cdot \mathbf{J}_\theta^{-1}$ can be found in the work of Salisbury [1] who assumed that the mechanical elasticity is concentrated in actuators and the deflections are small enough to apply linear approximation of the force-deflection relation. Here the matrix $\mathbf{K}_\theta$ aggregates the stiffness coefficients of all elastic joints, and $\mathbf{J}_\theta$ is the corresponding kinematic Jacobian. Further, this result was extended by Gosselin for the case of parallel manipulators taking into account elasticity of other mechanical elements [2]. More recent publications present VJM-based stiffness analysis for particular case studies, such as various variants of the Stewart–Gough platform, manipulators with US/UPS legs, CaPAMan, Orthoglide, H4 etc. [27-34].

It should be noted that in majority of related works, the presence of passive joints does not cause any specific computational problems, since these joints are eliminated via geometrical constraints describing assembling of the relevant parallel architecture [2]. Besides, in most of publications, it is implicitly assumed that the Jacobian $\mathbf{J}_\theta$ describing

Manuscript received March 7, 2011. The work presented in this paper was partially funded by the Region "Pays de la Loire", France and by the project ANR COROUSSO, France.

A. Pashkevich is with Ecole des Mines de Nantes, 4 rue Alfred-Kastler, Nantes 44307, France and with Institut de Recherches en Communications et en Cybernetique de Nantes, 44321 Nantes, France (phone: Tel. +33-251-85-83-00; fax. +33-251-85-83-49; e-mail: anatol.pashkevich@mines-nantes.fr).

A. Klimchik is with Ecole des Mines de Nantes, 4 rue Alfred-Kastler, Nantes 44307, France and with Institut de Recherches en Communications et en Cybernetique de Nantes, 44321 Nantes, France (e-mail: alexandr.klimchik@mines-nantes.fr).

S. Caro is with Institut de Recherches en Communications et en Cybernetique de Nantes, 44321 Nantes, France (e-mail: stephane.caro@irccyn.ec-nantes.fr).

D. Chablat is with Institut de Recherches en Communications et en Cybernetique de Nantes, 44321 Nantes, France (e-mail: damien.chablat@irccyn.ec-nantes.fr).

influence of the elastic joints on the end-location is non-singular[1], i.e. $rank(\mathbf{J}_\theta) = 6$, to ensure inversion of the related matrix in the modified expression $\mathbf{K}_C = (\mathbf{J}_\theta \cdot \mathbf{K}_\theta^{-T} \cdot \mathbf{J}_\theta^T)^{-1}$ that always produce non-singular $\mathbf{K}_C$. It is obvious that the assumption concerning $\mathbf{J}_\theta$ is completely realistic if the VJM model includes at least a single 6-dimensional virtual spring of a general type (see [35] for details), while it is not realistic that the manipulator stiffness matrix is always non-singular. Hence, common stiffness modelling techniques must be revised with respect to influence of passive joints, which in certain cases can not be straightforwardly eliminated from the kinetostatic equations and, consequently, may cause singularity of $\mathbf{K}_C$.

In this paper, it is applied another approach that originates from our publication [27] where the desired stiffness matrix $\mathbf{K}_C$ of size $6 \times 6$ is extracted from the inverse of a larger matrix, of size $(6 + n_q) \times (6 + n_q)$, which additionally includes the passive joint Jacobian $\mathbf{J}_q$ ($n_q$ is the passive joint number). Advantages of this approach and its ability to produce singular stiffness matrices were confirmed by a number of examples, but explicit analytical solution was not presented. Hence, this work concentrates on analytical computations of the stiffness matrix and also on influence of the passive joints on particular elements of $\mathbf{K}_C$.

It is also worth mentioning that some previous works [36] propose (or at least discuss) a trivial solution of the considered problem, which deals with a straightforward modification of the matrix $\mathbf{K}_\theta$, in accordance with the passive joint type and geometry (some rows and colons are simply set to zero). However, as it will be shown below, this straightforward approach gives true results if (and only if) the matrix $\mathbf{K}_\theta$ is diagonal, but it is not valid in general case where there is a coupling between different types of the elementary virtual springs presented by non-diagonal coefficients.

The remainder of this paper is organized as follows. Section 2 presents a simple motivating example that confirms the problem non-triviality. Then, Sections 3 and 4 propose relevant analytical solutions for a serial kinematic chain and a parallel robot respectively. Section 5 focuses on computational issues and proposes recursive procedure and a set of corresponding analytical rules. Section 6 contains application examples that demonstrate the developed technique advantages. And finally, Section 7 summarizes the main results and gives prospective for future work.

---

[1] It is important to distinguish the conventional kinematic Jacobian $\mathbf{J}$, which is computed with respect to actuated coordinates and may be both singular and non-singular, and the Jacobian $\mathbf{J}_\theta$ that is computed with respect to the virtual springs coordinates and is always non-singular. Besides, they differ in sizes, which for a standard serial 6-d.o.f. manipulator are respectively 6x6 and $6 \times 36$.

## II. MOTIVATING EXAMPLE

Let us present first a simple example that demonstrates non-trivial transformation of the stiffness matrix due to the presence of passive joints. For the purpose of simplicity, let us limit our study to 2D Cartesian space and consider a single manipulator link, which is assumed to be fixed at the left end. It is also assumed that the external loading (the forces $F_x$, $F_y$ and the torque $M_z$) is applied to the right end; either directly or via a passive joint.

Under these assumptions, the elastostatic properties of the link can be described by a symmetrical stiffness matrix $\mathbf{K} = [K_{ij}]$ of size $3 \times 3$ and its potential energy due to elastic deformations (linear deflections $\delta x$, $\delta y$ and angular deflection $\delta\varphi$) may be expressed as

$$E(\delta x, \delta y, \delta\varphi) = \frac{1}{2}[\delta x \ \delta y \ \delta\varphi] \begin{bmatrix} K_{11} & K_{12} & K_{13} \\ K_{21} & K_{22} & K_{23} \\ K_{31} & K_{32} & K_{33} \end{bmatrix} \begin{bmatrix} \delta x \\ \delta y \\ \delta\varphi \end{bmatrix} \quad (1)$$

If the link is equipped with a passive joint, the energy of this mechanical system (link with passive joint) must be minimised with respect to the joint variable. For instance, in the case of the rotational passive joint $R_z$ allowing free rotation around the z-axis at the reference point, the potential energy should be rewritten as

$$E_p(\delta x, \delta y) = \min_{\delta\varphi} E(\delta x, \delta y, \delta\varphi) \quad (2)$$

and the passive joint variable $\delta\varphi$ may be expressed via the remaining coordinates as $\delta\varphi = -(K_{13}\delta x + K_{23}\delta y)/K_{33}$. Then, after relevant transformations and computations of the second-order derivatives

$$K_{11}^p = \partial^2 E_p / \partial x^2, \ K_{12}^p = \partial^2 E_p / \partial x \partial y, ... K_{33}^p = \partial^2 E_p / \partial \varphi^2 \quad (3)$$

the desired stiffness matrix of links with passive joint may be expressed as

$$\mathbf{K}_p = \begin{bmatrix} K_{11} - \dfrac{K_{13} \cdot K_{31}}{K_{33}} & K_{12} - \dfrac{K_{32} \cdot K_{13}}{K_{33}} & 0 \\ K_{21} - \dfrac{K_{23} \cdot K_{31}}{K_{33}} & K_{22} - \dfrac{K_{23} \cdot K_{32}}{K_{33}} & 0 \\ \hdashline 0 & 0 & 0 \end{bmatrix} \quad (4)$$

This expression clearly shows that, if the matrix $\mathbf{K}$ is non-diagonal, a trivial transformation that was proposed in some previous works (i.e., simple setting to zero of the third raw and column) does not produce a truthful result. Moreover, the elements of the upper-left $2 \times 2$ block must be modified taking into account the elements of $\mathbf{K}$ that are located outside of this block. This conclusion motivates development of a general methodology of the stiffness matrix transformation, which is presented below.

## III. PASSIVE JOINTS IN A SERIAL CHAIN

In contrast to conventional serial manipulators, whose kinematics does not include passive joints and assures full controllability of the end-effector, parallel manipulators include a number of under-actuated serial chains that are mutually constrained by special connection to the base and to the end-platform. Let us derive an analytical expression for the stiffness matrix of such kinematic chain taking into account influence of the passive joints.

The kinematic chain under study (Fig.1) consists of a fixed base, a series of flexible links, a moving platform, and a number of actuated or passive joints separating these elements. Following the methodology proposed in our previous work [27], a relevant VJM model may be presented as a sequence of rigid links separated by passive joints and six-dimensional virtual springs describing elasticity of the links and actuators. For this VJM representation, the direct kinematics is defined by a product of homogeneous transformations that after extraction of the end-platform position and orientation is transformed into the vector function

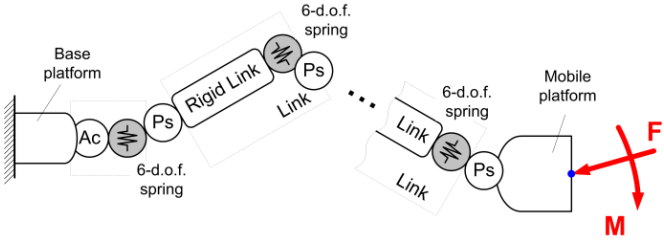

**Fig. 1.** The VJM model of a general serial chain
(Ps – passive joint, Ac – actuated joint)

$$\mathbf{t} = \mathbf{g}(\mathbf{q}, \boldsymbol{\theta}) \qquad (5)$$

where the vector $\mathbf{t} = (\mathbf{p}^T, \boldsymbol{\varphi}^T)^T$ includes the position $\mathbf{p} = (x, y, z)^T$ and orientation $\boldsymbol{\varphi} = (\varphi_x, \varphi_y, \varphi_z)^T$ of the platform in Cartesian space, the vector $\mathbf{q} = (q_1, q_2, ..., q_{n_q})^T$ contains passive joint coordinates, the vector $\boldsymbol{\theta} = (\theta_1, \theta_2, ..., \theta_{n_\theta})^T$ collects coordinates of all virtual springs; $n_q$ and $n_\theta$ are the sizes of $\mathbf{q}$ and $\boldsymbol{\theta}$ respectively.

It can be proved that the static equilibrium equations of this mechanical system may be written as

$$\mathbf{J}_\theta^T \cdot \mathbf{F} = \mathbf{K}_\theta \cdot \boldsymbol{\theta}; \qquad \mathbf{J}_q^T \cdot \mathbf{F} = \mathbf{0} \qquad (6)$$

where $\mathbf{J}_q = \partial \mathbf{g}(\mathbf{q}, \boldsymbol{\theta})/\partial \mathbf{q}$, $\mathbf{J}_\theta = \partial \mathbf{g}(\mathbf{q}, \boldsymbol{\theta})/\partial \boldsymbol{\theta}$ are kinematic Jacobians with respect to the passive and virtual joint coordinates respectively, $\mathbf{F}$ is the external loading (force and torque), and $\mathbf{K}_\theta$ the aggregated stiffness matrix of the virtual springs. Using these equations simultaneously with (5) and applying the first-order linear approximation under assumption that corresponding values of the external force $\mathbf{F}$ and the coordinate variations $\delta\mathbf{q}, \delta\boldsymbol{\theta}, \delta\mathbf{t}$ are small enough, one can derive the matrix expression

$$\begin{bmatrix} \mathbf{F} \\ \delta\mathbf{q} \end{bmatrix} = \begin{bmatrix} \mathbf{J}_\theta \cdot \mathbf{K}_\theta^{-1} \cdot \mathbf{J}_\theta^T & \mathbf{J}_q \\ \mathbf{J}_q^T & \mathbf{0} \end{bmatrix}^{-1} \cdot \begin{bmatrix} \delta\mathbf{t} \\ \mathbf{0} \end{bmatrix} \qquad (7)$$

that allows obtaining the desired Cartesian stiffness matrix $\mathbf{K}_C$ numerically. Corresponding procedure includes inversion of $(6+n_q) \times (6+n_q)$ matrix in the right-hand side of (7) and extracting from it the upper-left sub-matrix of the size $6 \times 6$ that defines a liner force-deflection relation in Cartesian space:

$$\mathbf{F} = \mathbf{K}_C \cdot \delta\mathbf{t} \qquad (8)$$

In spite of computational simplicity, the above procedure is not convenient for the parametric stiffness analysis that usually relies on analytical expressions. To derive such expression for the matrix $\mathbf{K}_C$, let us apply the blockwise inversion based on the Frobenius formula [37], that allows (after some transformations) to present the desired stiffness matrix as

$$\mathbf{K}_C = \mathbf{K}_C^0 - \mathbf{K}_C^0 \cdot \mathbf{J}_q \cdot \left( \mathbf{J}_q^T \cdot \mathbf{K}_C^0 \cdot \mathbf{J}_q \right)^{-1} \cdot \mathbf{J}_q^T \cdot \mathbf{K}_C^0 \qquad (9)$$

where the first term $\mathbf{K}_C^0 = (\mathbf{J}_\theta \cdot \mathbf{K}_\theta^{-1} \cdot \mathbf{J}_\theta^T)^{-1}$ is the stiffness matrix of the corresponding serial chain *without passive joints* and the second term defines the stiffness decrease *due to the passive joints*. It is worth mentioning that this result is in good agreement with other relevant works [14],[38],[40] where $\mathbf{K}_C$ was presented as the difference of two similar components but the second one was computed in a different way.

Analyzing the latter expression, one can get to the following conclusion concerning computational singularities:

**Remark 1.** The first term of the expression (9) is non-singular if and only if $rank(\mathbf{J}_\theta) = 6$, i.e. if the VJM model of the chain includes at least 6 independent virtual springs.

**Remark 2.** The second term of the expression (9) is non-singular if and only if $rank(\mathbf{J}_q) = n_q$, i.e. if the VJM model of the chain does not include redundant passive joints.

**Remark 3.** If both terms of (9) are non-singular, their difference produces a symmetrical stiffness matrix, which always singular and $rank(\mathbf{K}_C) = 6 - n_q$.

**Remark 4.** If the matrix $\mathbf{K}_C^0$ of the chain without passive joints is symmetrical and positive-definite, the stiffness matrix of the chain with passive joints $\mathbf{K}_C$ is also symmetrical but positive-semidefinite.

Hence, in practice, expression (5) does not cause any computational difficulties and always produce a singular stiffness matrix of rank $6 - n_q$. In analytical computations, the following proposition can be also useful that allows us to modify the original stiffness matrix $\mathbf{K}_C^0$ sequentially:

**Proposition.** If the chain does not include redundant passive joints, expression (5) allows recursive presentation

$$\mathbf{K}_C^{i+1} = \mathbf{K}_C^i - \mathbf{K}_C^i \cdot \mathbf{J}_q^i \cdot \left( \mathbf{J}_q^{i\,\mathrm{T}} \cdot \mathbf{K}_C^i \cdot \mathbf{J}_q^i \right)^{-1} \cdot \mathbf{J}_q^{i\,\mathrm{T}} \cdot \mathbf{K}_C^i; \quad i = 1, 2, \ldots \quad (10)$$

in which the sub-Jacobians $\mathbf{J}_q^i \subset \mathbf{J}_q$ are extracted from $\mathbf{J}_q = \left[ \mathbf{J}_q^1, \mathbf{J}_q^2, \ldots \right]$ in arbitrary order (column-by-column, or by groups of columns). Here $\mathbf{K}_C^0 = (\mathbf{J}_\theta \cdot \mathbf{K}_\theta^{-1} \cdot \mathbf{J}_\theta^\mathrm{T})^{-1}$

**Corollary**. The desired stiffness matrix $\mathbf{K}_C$ can be computed in $n_q$ steps, by sequential application of expression (6) for each single column of the Jacobian $\mathbf{J}_q$ (i.e. for each passive joint separately).

These results give convenient analytical and numerical computational techniques that are presented in details in Section 6.

## IV. PASSIVE JOINTS IN A PARALLEL MANIPULATOR

Let us consider now a parallel manipulator, which may be presented as a strictly parallel system of the actuated serial legs connecting the base and the end-platform (Fig. 2) [39]. Using the methodology described in previous section and applying it to each leg, there can be computed a set of $m$ Cartesian stiffness matrices $\mathbf{K}_C^{(i)}$, $i = 1, \ldots, m$ expressed with respect to the same coordinate system but corresponding to different platform points. If initially the chain stiffness matrices were computed in local coordinate systems, their transformation is performed in standard way [41], as

$$\mathbf{K}_C^{glob} = \begin{bmatrix} \mathbf{R} & \mathbf{0} \\ \mathbf{0} & \mathbf{R} \end{bmatrix} \cdot \mathbf{K}_C^{loc} \cdot \begin{bmatrix} \mathbf{R}^\mathrm{T} & \mathbf{0} \\ \mathbf{0} & \mathbf{R}^\mathrm{T} \end{bmatrix} \quad (11)$$

where $\mathbf{R}$ is a $3 \times 3$ rotation matrix describing orientation of the local coordinate system with respect to the global one.

To aggregate these matrices $\mathbf{K}_C^{(i)}$, they must be also re-computed with respect to same reference point of the platform. Assuming that the platform is rigid enough (compared to the legs), this conversion can be performed by extending the legs by a virtual rigid link connecting the end-point of the leg and the reference point of the platform (see Fig.2 where these extensions are defined by the vectors $\mathbf{v}_i$).

After such extension, an equivalent stiffness matrix of the leg may be expressed using relevant expression for a usual serial chain, i.e. as $\mathbf{J}_v^{(i)\,-\mathrm{T}} \cdot \mathbf{K}_C^{(i)} \cdot \mathbf{J}_v^{(i)\,-1}$, where the Jacobian $\mathbf{J}_v^{(i)}$ defines differential relation between the coordinates of the $i$-th virtual spring and the reference frame of the end-platform. Hence, the final expression for the stiffness matrix of the considered parallel manipulator can be written as

$$\mathbf{K}_C^{(m)} = \sum_{i=1}^{m} \mathbf{J}_v^{(i)\,-\mathrm{T}} \cdot \mathbf{K}_C^{(i)} \cdot \mathbf{J}_v^{(i)\,-1} \quad (12)$$

where $m$ is the number of serial kinematic chains in the manipulator architecture. Besides, it is implicitly assumed here that all stiffness matrices (both for the legs and for the whole manipulator) are expressed in the same global coordinate system. Hence, the axes of all virtual springs are parallel to the axes $x$, $y$, $z$ of this system and corresponding Jacobians and their inverses can be easily computed analytically as

$$\mathbf{J}_v^{(i)} = \begin{bmatrix} \mathbf{I}_3 & (\mathbf{v}_i \times) \\ \mathbf{0} & \mathbf{I}_3 \end{bmatrix}_{6 \times 6}, \quad \mathbf{J}_v^{(i)-1} = \begin{bmatrix} \mathbf{I}_3 & -(\mathbf{v}_i \times) \\ \mathbf{0} & \mathbf{I}_3 \end{bmatrix}_{6 \times 6} \quad (13)$$

where $\mathbf{I}_3$ is a identity matrix of size $3 \times 3$, $\mathbf{v}_i$ is the vector from the leg end-point to the platform reference point (see Fig.2) and $(\mathbf{v} \times)$ is a skew-symmetric matrix corresponding to the vector $\mathbf{v}$:

$$(\mathbf{v} \times) = \begin{bmatrix} 0 & -v_z & v_y \\ v_z & 0 & -v_x \\ -v_y & v_x & 0 \end{bmatrix} \quad (14)$$

Therefore, expression (12) allows explicit aggregation of the leg stiffness matrices with respect to any given reference point of the platform. It is worth mentioning that in practice, the matrices $\mathbf{K}_C^{(i)}$ are always singular while there aggregation usually produce non-singular singular matrix. Relevant examples are presented in following sections.

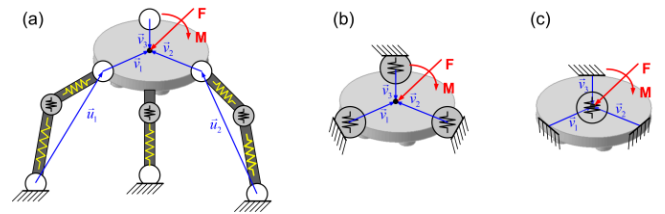

**Fig. 2.** Typical parallel manipulator (a) and transformation of its VJM models (b, c)

## V. COMPUTATIONAL TECHNIQUES

Explicit expressions (9), (11) derived in previous sections allow obtaining the Cartesian stiffness matrix instantly, for any Jacobian $\mathbf{J}_q$ describing special location of the passive joints. However, recursive equation (10) allows essentially *simplify the computational procedure* by sequential modification of the original stiffness matrix $\mathbf{K}_C^0$ for each passive joint independently, using separate columns of $\mathbf{J}_q = [\mathbf{J}_{q1}, \mathbf{J}_{q2}, ...]$. Moreover, for some typical cases, relevant computations may be easily performed analytically. This section presents some useful techniques related to this approach.

### A. Recursive computations: single-joint decomposition

Let us assume that a current recursion deals with a single passive joint corresponding to the *i*-th column of the Jacobian $\mathbf{J}_q$, which is denoted as $\mathbf{J}_q^i$ and has size $6 \times 1$. In this case, the matrix expression $(\mathbf{J}_q^{i\,T} \cdot \mathbf{K}_C^i \cdot \mathbf{J}_q^i)^{-1}$ is reduced to the size of $1 \times 1$ and the matrix inversion is replaced by a simple scalar division. Besides, the term $\mathbf{K}_C^i \cdot \mathbf{J}_q^i$ has size $6 \times 1$, so the recursion (10) is simplified to

$$\mathbf{K}_C^{i+1} = \mathbf{K}_C^i - \frac{1}{\mu} \mathbf{u}_i \cdot \mathbf{u}_i^T \quad or \quad \left[ K_{jk}^{(i+1)} \right] = \left[ K_{jk}^{(i)} \right] - \frac{1}{\mu} \left[ u_j^{(i)} u_k^{(i)} \right] \quad (15)$$

where $\mathbf{u}_i = \mathbf{K}_C^i \cdot \mathbf{J}_q^i$ is a $6 \times 1$ vector and $\mu = \mathbf{J}_q^{i\,T} \cdot \mathbf{K}_C^i \cdot \mathbf{J}_q^i$ is a scalar. It can be also proved that each recursions reduces the rank of the stiffness matrix by 1

$$rank\left(\mathbf{K}_C^{i+1}\right) = rank\left(\mathbf{K}_C^i\right) - 1 \quad (16)$$

provided that the current Jacobian $\mathbf{J}_q^i$ is independent of the previous ones $\mathbf{J}_q^1, \mathbf{J}_q^2 ...$ (i.e. the *i*-th passive joints is not redundant relatively to the joints $1, ..., i-1$).

Since in practice any combination of passive joints can be decomposed into elementary translational and rotational ones, it is enough to consider only two types of the Jacobian columns $\mathbf{J}_{qi}$:

$$\begin{aligned}\mathbf{J}_{tran} &= \begin{bmatrix} e_1 & e_2 & e_3 & 0 & 0 & 0 \end{bmatrix}^T; \\ \mathbf{J}_{tran} &= \begin{bmatrix} d_1 & d_2 & d_3 & e_1 & e_2 & e_3 \end{bmatrix}^T \end{aligned} \quad (17)$$

where the unit vector $\mathbf{e} = [e_1 \ e_2 \ e_3]$, $\mathbf{e}^T\mathbf{e} = 1$ defines orientation of the passive joint axis (both for translational and rotational ones) and the vector $\mathbf{d} = [d_1 \ d_2 \ d_3]$ defines influence of the rotational passive joints on the linear velocity at the reference point, i.e. $\mathbf{d} = \mathbf{e} \times \mathbf{r}$ where $\mathbf{r}$ is a vector from the joint centre point to the reference point.

Hence, in general case, the recursion (10) involves rather intricate matrix transformation, different from simple setting to zero a row and/or a column. Let us consider now several specific (but rather typical) cases where the transformation rules are more simple and elegant.

### B. Analytical computations: trivial passive joints

In practice, many parallel robots include kinematic chains for which the passive joint axes are collinear to the axes x, y or z of the Cartesian coordinate system. For such architectures, the vector-columns of the Jacobian $\mathbf{J}_q$ include a number of zero elements, so the expressions (13) can be essentially simplified. Let us consider a set of trivial cases where $\mathbf{J}_q^i$ are created from the columns of the identity matrix $\mathbf{I}_{6 \times 6}$:

Corresponding passive joints will be further referred to as the 'trivial' ones. It can be easily proved that they cover the following range of the joint geometry:

(i) translational passive joint with arbitrary spatial position (but with the joint axis directed along *x*, *y* or *z*);
(ii) rotational passive joints positioned at the reference point (and with the joint axis directed along *x*, *y* or *z*).

Besides, it is worth to consider additional case-study corresponding to

(iii) rotational passive joints shifted by a distance *L* with respect to the reference point in the direction either *x*, *y* or *z* (and with the joint axis directed along *x*, *y* or *z*), which will be further referred to as the 'quasi-trivial' and gives the Jacobian columns of the following structure:

$$\mathbf{J}_q^{(4+)} = \begin{bmatrix} 0 \\ d_y \\ d_z \\ 1 \\ 0 \\ 0 \end{bmatrix}, \quad \mathbf{J}_q^{(5+)} = \begin{bmatrix} d_x \\ 0 \\ d_z \\ 0 \\ 1 \\ 0 \end{bmatrix}, \quad \mathbf{J}_q^{(6+)} = \begin{bmatrix} d_x \\ d_y \\ 0 \\ 0 \\ 0 \\ 1 \end{bmatrix} \quad (18)$$

where $d_x, d_y, d_z$ denote the elements of the vector $\mathbf{d}$, which are equal here either $\pm L$ or 0.

For the trivial passive joints, assuming that $\mathbf{J}_q^{(p)}$ denotes the vector-column with a single non-zero element in the *p*-th position, a straightforward substitution yields $u_j = K_{jp}^{(i)}$; $\mu = K_{pp}^{(i)}$. So, the recursive expression (9) for the Cartesian stiffness matrix is simplified to

$$\left[ K_{jk}^{(i+1)} \right] = \left[ K_{jk}^{(i)} \right] - \left[ \frac{K_{jk}^{(i)} K_{kj}^{(i)}}{K_{pp}^{(i)}} \right] \quad (19)$$

that is very similar to those presented in the motivating example (see Section 2). Also, here the *p*–th row and column

of the matrix $\mathbf{K}_C^{i+1}$ become equal to zero

$$K_{pk}^{(i+1)} = 0; \quad K_{kp}^{(i+1)} = 0; \quad \forall p = 1,...6 \quad (20)$$

and the recursive computations are easily performed analytically.

## VI. APPLICATION EXAMPLES

Let apply now the developed technique to computing of the stiffness matrix for two versions of a general Stewart-Gough platform presented in Fig. 3 [42]-[44]. It is assumed that in both cases the manipulator base and the moving plate (platform) are connected by six similar extensible legs (Fig. 4) but their spatial arrangements are different:

**Case A**: the legs are regularly connected to the base and platform, with the same angular distance 60° (it is obviously a degenerate design, where the stiffness matrix should be singular)

**Case B**: the legs are connected to the base and platform in three pairs, with the angular distance of 120° between the mounting points (it is a classical design of Stewart-Gough where the stiffness matrix should be non-singular).

For both designs, the original leg stiffness (i.e. without the passive joints) can be described by the sparse matrix corresponding to the symmetric beam. Further, to take into account the passive joints influence, the procedure (10) should be applied recursively, using the elementary Jacobians

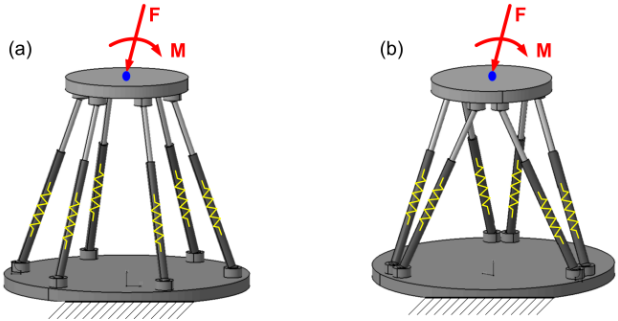

**Fig. 3** Geometry of the Stewart-Gough platforms under study

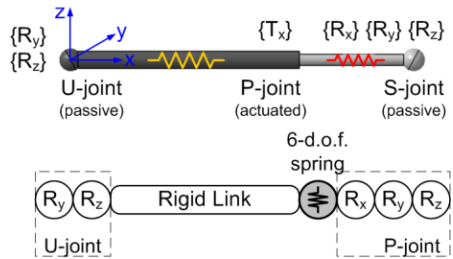

**Fig. 4.** Geometry of the manipulator leg and its VJM model

$$\mathbf{J}_{Rx}^{(1)} = \begin{bmatrix} 0 \\ 0 \\ 0 \\ 1 \\ 0 \\ 0 \end{bmatrix}; \mathbf{J}_{Ry}^{(2)} = \begin{bmatrix} 0 \\ 0 \\ 0 \\ 0 \\ 1 \\ 0 \end{bmatrix}; \mathbf{J}_{Rz}^{(3)} = \begin{bmatrix} 0 \\ 0 \\ 0 \\ 0 \\ 0 \\ 1 \end{bmatrix}; \mathbf{J}_{Ry+}^{(4)} = \begin{bmatrix} 0 \\ 0 \\ -L \\ 0 \\ 1 \\ 0 \end{bmatrix}; \mathbf{J}_{Rz+}^{(5)} = \begin{bmatrix} 0 \\ L \\ 0 \\ 0 \\ 0 \\ 1 \end{bmatrix}$$

(21)

where $L$ is the leg length. It is obvious that, due to trivial structure of $\mathbf{J}_q^i$, the recursive computations can be easily performed analytically.

Hence, in final form, the derived matrix includes only the traction/compression term (and not bending, torsion, etc.) what perfectly agrees with other results on Stewart-Gough platforms.

Further, to be applied to each leg, the obtained matrix must be transformed from the local to the global coordinate system. In this specific case, due to the special structure of $\mathbf{K}_C^5$, relevant transformation [24][41]

$$\mathbf{K}_{C_i} = \begin{bmatrix} \mathbf{R}_i & 0 \\ 0 & \mathbf{R}_i \end{bmatrix} \cdot K_{11} \cdot \begin{bmatrix} 1 & \mathbf{0}_{1\times 5} \\ \mathbf{0}_{5\times 1} & \mathbf{0}_{5\times 5} \end{bmatrix} \cdot \begin{bmatrix} \mathbf{R}_i^T & 0 \\ 0 & \mathbf{R}_i^T \end{bmatrix} \quad (22)$$

expressed via the orthogonal rotation matrix $\mathbf{R}_i$ describing orientation of the $i$-th local coordinate system with respect to the global one, is easily reduced to

$$\mathbf{K}_{C_i} = K_{11} \cdot \begin{bmatrix} \mathbf{u}_i^0 \cdot \mathbf{u}_i^{0T} & \mathbf{0}_{3\times 3} \\ \mathbf{0}_{3\times 3} & \mathbf{0}_{3\times 3} \end{bmatrix}. \quad (23)$$

where $\mathbf{u}_i^0$ is the unit vector directed along the leg axis $\mathbf{u}_i$ (see Fig.2). Besides, before aggregation, the stiffness matrices of separate legs $\mathbf{K}_{C_i}$ must be re-computed with respect to same reference point in accordance with expressions (12), (13) which yields

$$\mathbf{K}_C = K_{11} \sum_{i=1}^{6} \begin{bmatrix} \mathbf{I}_3 & 0 \\ (\mathbf{v}_i \times) & \mathbf{I}_3 \end{bmatrix} \begin{bmatrix} \mathbf{u}_i^0 \cdot \mathbf{u}_i^{0T} & \mathbf{0}_{3\times 3} \\ \mathbf{0}_{3\times 3} & \mathbf{0}_{3\times 3} \end{bmatrix} \begin{bmatrix} \mathbf{I}_3 & (\mathbf{v}_i \times)^T \\ 0 & \mathbf{I}_3 \end{bmatrix}$$

(24)

where $\mathbf{v}_i$ is the vector from the leg end-point to the platform reference point (see Fig.2). So, after relevant transformations, one can get the final expression of the manipulator stiffness matrix

$$\mathbf{K}_C = K_{11} \cdot \sum_{i=1}^{6} \begin{bmatrix} \mathbf{u}_i^0 \\ (\mathbf{v}_i \times \mathbf{u}_i^0) \end{bmatrix} \cdot \begin{bmatrix} \mathbf{u}_i^{0T} & (\mathbf{v}_i \times \mathbf{u}_i^0)^T \end{bmatrix} \quad (25)$$

where the vector $\mathbf{u}_i$, $\mathbf{v}_i$ describing spatial locations of the legs and computed via the direct kinematics, and $\mathbf{v}_i \times \mathbf{u}_i^0$ denotes the vector product which is referred to the corresponding skew-symmetric matrix).

The derived equation was applied to both case studies, assuming that the manipulators are in their "*home*" configurations when the platform is parallel to the base and it is symmetrical with respect to the vertical axis. Corresponding expressions for the leg vectors are

$$\mathbf{u}_i = \begin{bmatrix} r\cos(\psi_i) - R\cos(\phi_i) \\ r\sin(\psi_i) - R\sin(\phi_i) \\ h \end{bmatrix}, \quad \mathbf{v}_i = \begin{bmatrix} -r\cos(\psi_i) \\ -r\sin(\psi_i) \\ 0 \end{bmatrix} \quad (26)$$

where for the case A $\phi_i = \psi_i \in \{0, 60°, 120°, 180°, 240°, 300°\}$, and for the case B $\phi_i \in \{0, 120°, 120°, 240°, 240°, 360°\}$; $\psi_i \in \{60, 60, 180°, 180°, 300°, 300°\}$, $h$ is the vertical distance between the base and the platform. Substitution of these vectors to the expression (25) leads to the following stiffness matrices

$$\mathbf{K}_C^{(A)} = \frac{3K_{11}}{L^2} \begin{bmatrix} d_a^2 & 0 & 0 & 0 & rhd_a & 0 \\ 0 & d_a^2 & 0 & -rhd_a & 0 & 0 \\ 0 & 0 & 2h^2 & 0 & 0 & 0 \\ \hdashline 0 & -rhd_a & 0 & r^2h^2 & 0 & 0 \\ rhd_a & 0 & 0 & 0 & r^2h^2 & 0 \\ 0 & 0 & 0 & 0 & 0 & 0 \end{bmatrix} \quad (27)$$

and

$$\mathbf{K}_C^{(B)} = \frac{3K_{11}}{L^2} \begin{bmatrix} d_a^2 + Rr & 0 & 0 & 0 & rhd_b & 0 \\ 0 & d_a^2 + Rr & 0 & -rhd_b & 0 & 0 \\ 0 & 0 & 2h^2 & 0 & 0 & 0 \\ \hdashline 0 & -rhd_b & 0 & r^2h^2 & 0 & 0 \\ rhd_b & 0 & 0 & 0 & r^2h^2 & 0 \\ 0 & 0 & 0 & 0 & 0 & 1.5\,r^2R^2 \end{bmatrix}$$

(28)

where $R$ and $r$ denote the circle radius which comprise the leg connection point at the base and moving platform respectively, $d_a = R - r$; $d_b = R/2 - r$; $L$ is the leg length, and the superscripts 'A' and 'B' define the relevant case study. As follows from these expressions, in "home" location, the matrix $\mathbf{K}_C^{(A)}$ is singular and allows "free" rotation of the end-platform around the vertical axis. In contrast, for the same location, the matrix $\mathbf{K}_C^{(B)}$ is non-singular and the manipulator resists to all external forces/torques applied to the platform. These results are in good agreement with previous research on the Stewart-Gough platforms and confirm efficiency of the developed computational technique for manipulator stiffness modeling.

## VII. CONCLUSION

For robotic manipulators with passive joints, the stiffness matrices of separate kinematic chains are *singular*. So, the most of existing stiffness analysis methods can not be applied directly and this problem is usually solved by elimination the passive joint coordinates via geometrical constraints describing the manipulator assembly. However, such techniques degenerate if the number of passive joints is redundant and/or the resulting matrix is inherently singular.

To deal with such architectures in more efficient way, this paper proposes an analytical approach that allows obtaining both *singular and non-singular* stiffness matrices and which is appropriate for a general case, independent of the type and spatial location of the passive joints. The developed approach is based on the extension of the virtual-joint modelling technique and includes two basic steps which sequentially produce stiffness matrices of separate chains and then aggregate them in a common matrix.

In contrast to previous works, the desired stiffness matrix is presented in an explicit *analytical form*, as a sum of *two terms*. The first of them has traditional structure and describes manipulator elasticity due to the link/joint flexibility, while the second one directly takes into account influence of the passive joints. It is proved that, for each chain, the rank-deficiency of the resulting matrix is equal to the number of independent passive joints. To simplify analytical computations, it is proposed a *recursive procedure* that sequentially modifies the original matrix in accordance with the geometry of each passive joint. For the trivial cases, for which the passive joint axes are collinear to the axes of the base coordinate system, this modification is presented in the form of simple analytical rules.

Advantages of the developed technique are illustrated by application examples that deal with stiffness modelling of two Stewart-Gough platforms. They demonstrate its ability to produce both singular and non-singular stiffness matrices, and also show its feasibility for analytical computations. These examples give also some prospective for future work that include development of the dedicated techniques for the stiffness matrix aggregation in the case of non-rigid platform and an extension of these results for the case of manipulators with external loading.

## ACKNOWLEDGEMENTS

The work presented in this paper was partially funded by the Region "Pays de la Loire" (France), by the EU commission (project NEXT) and by the ANR, France (project COROUSSO).

# REFERENCES

[1] J. Salisbury, Active Stiffness Control of a Manipulator in Cartesian Coordinates, in: 19th IEEE Conference on Decision and Control, 1980, pp. 87–97.

[2] C. Gosselin, Stiffness mapping for parallel manipulators, IEEE Transactions on Robotics and Automation 6(3) (1990) 377–382.

[3] B.-J. Yi, R.A. Freeman, Geometric analysis antagonistic stiffness redundantly actuated parallel mechanism, Journal of Robotic Systems 10(5) (1993) 581-603.

[4] Griffis, M., Duffy, J.: Global stiffness modeling of a class of simple compliant couplings. Mechanism and Machine Theory **28**(2), 207–224 (1993)

[5] Ciblak, N., Lipkin, H.: Asymmetric Cartesian stiffness for the modeling of compliant robotic systems. In: Proc. 23rd Biennial ASME Mechanisms Conference, Minneapolis, MN (1994)

[6] T. Pigoski, M. Griffis, J. Duffy, Stiffness mappings employing different frames of reference. Mechanism and Machine Theory 33(6) (1998) 825–838.

[7] Howard, S., Zefran, M., Kumar, V.: On the 6 x 6 Cartesian stiffness matrix for three-dimensional motions. Mechanism and Machine Theory **33**(4), 389–408 (1998)

[8] S. Chen, I. Kao, Conservative Congruence Transformation for Joint and Cartesian Stiffness Matrices of Robotic Hands and Fingers, The International Journal of Robotics Research 19(9) (2000) 835–847

[9] M.M. Svinin, S. Nosoe, M. Uchiyama, On the stiffness and stability of Gough-Stewart platforms, in: Proc. of IEEE International Conference on Robotics and Automation (ICRA), 2001, pp. 3268–3273

[10] C.M. Gosselin, D. Zhang, Stiffness analysis of parallel mechanisms using a lumped model, International Jornal of Robotics and Automation 17 (2002) 17-27.

[11] O. Company, F. Pierrot, J.-C. Fauroux, A Method for Modeling Analytical Stiffness of a Lower Mobility Parallel Manipulator, in: Proceedings of IEEE International Conference on Robotics and Automation (ICRA), 2005, pp. 3232 - 3237

[12] G. Alici, B. Shirinzadeh, Enhanced stiffness modeling, identification and characterization for robot manipulators, Proceedings of IEEE Transactions on Robotics 21(4) (2005) 554–564.

[13] J. Kövecses, J. Angeles, The stiffness matrix in elastically articulated rigid-body systems, Multibody System Dynamics 18(2) (2007) 169–184.

[14] C. Quennouelle, C. M. Gosselin, Stiffness Matrix of Compliant Parallel Mechanisms, In: Springer Advances in Robot Kinematics: Analysis and Design, 2008, pp. 331-341.

[15] J.-P. Merlet, C. Gosselin, Parallel mechanisms and robots, In B. Siciliano, O. Khatib, (Eds.), Handbook of robotics, Springer, Berlin, 2008, pp. 269-285.

[16] J.-P. Merlet, Analysis of Wire Elasticity for Wire-driven Parallel Robots, In: Proceedings of the Second European Conference on Mechanism Science (EUCOMES 08), Springer, 2008, pp. 471-478

[17] T. Bonnemains, H. Chanel, C. Bouzgarrou and P. Ray, Definition of a new static model of parallel kinematic machines: highlighting of overconstraint influence, in: Proceedings of IEEE Int. Conference on Intelligent Robots and Systems (IROS), 2008, pp. 2416–2421.

[18] I. Tyapin, G. Hovland, Kinematic and elastostatic design optimization of the 3-DOF Gantry-Tau parallel kinamatic manipulator,Modelling, Identification and Control, 30(2) (2009) 39-56

[19] G. Piras, W.L. Cleghorn, J.K. Mills, Dynamic finite-element analysis of a planar high-speed, high-precision parallel manipulator with flexible links, Mechanism and Machine Theory 40 (7) (2005) 849–862.

[20] X. Hu, R. Wang, F. Wu, D. Jin, X. Jia, J. Zhang, F. Cai, Sh. Zheng, Finite Element Analysis of a Six-Component Force Sensor for the Trans-Femoral Prosthesis, In: V.G. Duffy (Ed.), Digital Human Modeling, Springer-Verlag, Berlin Heidelberg, 2007, pp. 633–639.

[21] K. Nagai, Zh. Liu, A Systematic Approach to Stiffness Analysis of Parallel Mechanisms and its Comparison with FEM, In: Proceeding of SICE Annual Conference, Japan, 2007, pp 1087-1094.

[22] B.C. Bouzgarrou, J.C. Fauroux, G. Gogu, Y. Heerah, Rigidity analysis of T3R1 parallel robot with uncoupled kinematics, In: Proceedings Of the35th International Symposium on Robotics, 2004.

[23] R. Rizk, J.C. Fauroux, M. Mumteanu, G. Gogu, A comparative stiffness analysis of a reconfigurable parallel machine with three or four degrees of mobility, Journal of Machine Engineering 6 (2) (2006) 45–55.

[24] D. Deblaise, X. Hernot, P.Maurine, A systematic analytical method for PKM stiffness matrix calculation, In: Proceedings of the IEEE International Conference on Robotics and Automation (ICRA), Orlando, Florida, 2006, pp. 4213-4219.

[25] H. C. Martin, Introduction to matrix methods of structural analysis, McGraw-Hill Education, 1966

[26] Sh.-F. Chen, I. Kao, Geometrical Approach to The Conservative Congruence Transformation (CCT) for Robotic Stiffness Control, In: Proceedings of the 2002 IEEE lntemational Conference on Robotics and Automation (ICRA) Washington, DC, 2002, pp 544-549.

[27] A. Pashkevich, D. Chablat, P. Wenger, Stiffness analysis of overconstrained parallel manipulators, Mechanism and Machine Theory 44 (2009) 966-982.

[28] M. Ceccarelli, G. Carbone, A stiffness analysis for CaPaMan (Cassino Parallel Manipulator), Mechanism and Machine Theory 37 (5) (2002) 427–439.

[29] O. Company, S. Krut, F. Pierrot, Modelling and preliminary design issues of a 4-axis parallel machine for heavy parts handling, Journal of Multibody Dynamics 216 (2002) 1–11.

[30] D. Zhang, F. Xi, C.M. Mechefske, S.Y.T. Lang, Analysis of parallel kinematic machine with kinetostatic modeling method, Robotics and Computer-Integrated Manufacturing 20 (2) (2004) 151–165.

[31] R. Vertechy, V. Parenti-Castelli, Static and stiffness analyses of a class of over-constrained parallel manipulators with legs of type US and UPS, in: Proceedings of IEEE International Conference on Robotics and Automation (ICRA), 2007, pp. 561–567.

[32] Y. Li, Q. Xu, Stiffness analysis for a 3-PUU parallel kinematic machine, Mechanism and Machine Theory 43(2) (2008) 186-200.

[33] F. Majou, C. Gosselin, P. Wenger, D. Chablat, Parametric stiffness analysis of the Orthoglide, Mechanism and Machine Theory 42 (2007) 296-311.

[34] O. Company, S. Krut, F. Pierrot, Modelling and preliminary design issues of a 4-axis parallel machine for heavy parts handling, Journal of Multibody Dynamics 216 (2002) 1–11.

[35] A. Pashkevich, A. Klimchik, D. Chablat, Ph. Wenger, Accuracy Improvement for Stiffness Modeling of Parallel Manipulators, In: Proceedings of 42nd CIRP Conference on Manufacturing Systems, Grenoble, France, 2009.

[36] A. Taghaeipour, J. Angeles, L. Lessard, Online computation of the stiffness matrix in robotic structures using finite element analysis, report TR-CIM-10-05, Department of Mechanical engineering and centre for intelligent machines, McGill university, 2010

[37] F. Gantmacher, Theory of matrices, AMS Chelsea publishing, 1959

[38] J. Chen, F. Lan, Instantaneous stiffness analysis and simulation for hexapod machines, Simulation Modelling Practice and Theory 16 (2008) 419–428

[39] J.-P. Merlet, Parallel Robots, Kluwer Academic Publishers, Dordrecht, 2006.

[40] C. Quennouelle, C. M.Gosselin, Instantaneous Kinemato-Static Model of Planar Compliant Parallel Mechanisms, In: Proceedings of ASME International Design Engineering Technical Conferences, Brooklyn, NY, USA, 2008.

[41] J. Angeles, Fundamentals of Robotic Mechanical Systems: Theory, Methods, and Algorithms, Springer, New York, 2007.

[42] Y.W. Li, J.S. Wang, L.P. Wang (2002). Stiffness analysis of a Stewart platform-based parallel kinematic machine, In: Proceedings of IEEE International Conference on Robotics and Automation (ICRA), Washington, US, 2002(4), pp. 3672–3677.

[43] B.S. El-Khasawneh, P.M. Ferreira, Computation of stiffness and stiffness bounds for parallel link manipulators, International Journal of Machine Tools and Manufacture 39 (2) (1999) 321–342.

[44] H.K. Arumugam, R.M. Voyles, S. Bapat, Stiffness analysis of a class of parallel mechanisms for micro-positioning applications, in: Proceedings of IEEE/ RSJ International Conference on Intelligent Robots and Systems (IROS), 2004, vol. 2, pp. 1826–1831.

[45] G. Strang, Introduction to Linear Algebra, Wellesley, MA, Wellesley Cambridge Press, 1998.